\documentclass{article}
\usepackage{spconf,amsmath,graphicx}
\usepackage{amsmath,graphicx,amsfonts,epsfig,epstopdf,caption,subcaption,mathrsfs}

\usepackage{color}

\title{Recurrent Neural Network Training with Dark Knowledge Transfer}
\name{Zhiyuan Tang$^{1,3}$, Dong Wang$^{1,2*}$, Zhiyong Zhang$^{1,2}$\thanks{{This work was supported by the National Natural Science Foundation of China under Grant No. 61371136  and the MESTDC PhD Foundation Project No. 20130002120011. This paper was also supported by Huilan Ltd. and Sinovoice.}}}
\address{1. Center for Speech and Language Technologies (CSLT), RIIT, Tsinghua University \\
2. Tsinghua National Laboratory for Information Science and Technology \\
3. Chengdu Institute of Computer Applications, Chinese Academy of Sciences \\
{\small \tt \{tangzy,zhangzy\}@cslt.riit.tsinghua.edu.cn} \\
{\small \tt $^*$Corresponding Author:wangdong99@mails.tsinghua.edu.cn}
}

\begin{document}
%
\maketitle
\begin{abstract}

Recurrent neural networks (RNNs), particularly long short-term memory (LSTM), have gained much attention in automatic speech recognition (ASR). Although some successful stories have been reported, training RNNs remains highly challenging, especially with limited training data.
Recent research found that a well-trained model can be used as a teacher to train other child models, by using the predictions generated by the teacher model as supervision. This knowledge transfer learning has been employed to train simple neural nets with a complex one, so that the final performance can reach a level that is infeasible to obtain by regular training. In this paper, we employ the knowledge transfer learning approach to train RNNs (precisely LSTM) using a deep neural network (DNN) model as the teacher. This is different from most of the existing research on knowledge transfer learning, since the teacher (DNN) is assumed to be weaker than the child (RNN); however, our experiments on an ASR task showed that it works fairly well: without applying any tricks on the learning scheme, this approach can train RNNs successfully even with limited training data.

\end{abstract}
\section{Introduction}

Deep learning has gained significant success in a wide range of applications, for example, automatic speech recognition (ASR)~\cite{deng2014}. A powerful deep learning model that has been reported effective in ASR is the recurrent neural network (RNN), e.g., \cite{graves2013speech,graves2014towards,sak2014long}. An obvious advantage of RNNs compared to conventional deep neural networks (DNNs) is that RNNs can model long-term temporal properties and thus are suitable for modeling speech signals.

A simple training method for RNNs is the backpropagation through time algorithm~\cite{rumelhart1988learning}. This first-order approach, however, is rather inefficient due to two main reasons: (1) the twists of the objective function caused by the high nonlinearity; (2) the vanishing and explosion of gradients in backpropagation~\cite{bengio1994learning}. In order to address these difficulties (mainly the second), a modified  architecture  called  the  long  short-term memory (LSTM) was proposed in~\cite{hochreiter1997long} and has been successfully applied to ASR~\cite{graves2005framewise}. In the echo state network (ESN) architecture proposed by~\cite{jaeger2004harnessing}, the hidden-to-hidden weights are not learned in the training so the problem of odd gradients does not exist. Recently, a special variant of the Hessian-free (HF) optimization approach was successfully applied to learn RNNs from random initialization~\cite{martens2010deep,martens2011learning}. A particular problem of the HF approach is that the computation is demanding. Another recent study shows that a carefully designed momentum setting can significantly improve RNN training, with limited computation and can reach the performance of the HF method~\cite{sutskever2013importance}. Although these methods can address the difficulties of RNN training to some extent, they are either too tricky (e.g., the momentum method) or less optimal (e.g., the ESN method). Particularly with limited data, RNN training remains difficult.

This paper focuses  on the LSTM structure and presents a simple yet powerful training algorithm based on knowledge transfer. This algorithm is largely motivated by the recently proposed logit matching~\cite{ba2014deep} and dark knowledge distiller~\cite{hinton2014distilling}. The basic idea of the knowledge transfer approach is that a well-trained model involves rich knowledge of the target task and can be used to guide the training of other models. Current research focuses on learning simple models (in terms of structure) from a powerful yet complex model, or an ensemble of models~\cite{ba2014deep,hinton2014distilling} based on the idea of model compression~\cite{bucilu¨£2006model}. In ASR, this idea has been employed to train small DNN models from a large and complex one~\cite{li2014learning}.

In this paper, we conduct an opposite study, which employs a simple DNN model to train a more complex RNN. Different from the existing research that tries to distill knowledge from the teacher model, we treat the teacher model as a regularization so that the training process of the child model is smoothed, or a pre-training step so that the supervised training can be located at a good starting point. This in fact leads to a new training approach that is easy to perform and can be extended to any model architecture. We employ this idea to address the difficulties in RNN training. The experiments on an ASR task with the Aurora4 database verified that the proposed method can significantly improve RNN training.

The reset of the paper is organized as follows. Section~\ref{sec:rel} briefly discusses some related works, and Section~\ref{sec:method} presents the method. Section~\ref{sec:exp} presents the experiments, and the paper is concluded by Section~\ref{sec:con}.

\section{Related to prior work}
\label{sec:rel}

This study is directly motivated by the work of dark knowledge distillation~\cite{hinton2014distilling}. The important aspect that distinguishes our work from others is that the existing methods focus on distilling knowledge of complex model and use it to improve simple models, whereas our study uses simple models to teach complex models.  The teacher model in our work in fact knows not so much, but it is sufficient to provide a rough guide that is important to train complex models, such as RNNs in the present study.

Another related work is the knowledge transfer between DNNs and RNNs, as proposed in~\cite{chan2015transferring}. However, it employs knowledge transfer to train DNNs with RNNs. This still follows the conventional idea described above, and so is different from ours.

\section{RNN training with knowledge transfer}
\label{sec:method}

\subsection{Dark knowledge distiller}

The idea that a well-trained DNN model can be used as a teacher to guide the training of other models was proposed by several authors almost at the same time~\cite{ba2014deep,hinton2014distilling,li2014learning}. The basic assumption is that the teacher model encodes rich knowledge for the task in hand and this knowledge can be distilled to boost the child model which is often simpler and can not learn many details without the teacher's guide. There are a few ways to distill the knowledge. The logit matching approach proposed by~\cite{ba2014deep} teaches a child model by encouraging its logits (activations before softmax) close  to those of the teacher model in terms of the $\ell$-2 norm, and the dark knowledge distiller model proposed by~\cite{hinton2014distilling} encourages the posterior probabilities (softmax output) of the child model close to those of the teacher model in terms of cross entropy. This transfer learning has been applied to learn simple models to approach the performance of a complex model or a large model ensemble, for example, learning a small DNN from a large DNN~\cite{li2014learning} or a DNN from a more complex RNN~\cite{chan2015transferring}.

We focus on the dark knowledge distiller approach as it showed better performance in our experiments. Basically, a well-trained DNN model plays the role of a teacher and generates posterior probabilities of the training samples as new targets for training other models. These posterior probabilities are called `soft targets'  since the class identities are not as deterministic as the original one-hot `hard targets'. To make the targets softer, a temperature $T$ can be applied to scale the logits in the softmax, formulated as $p_i = \frac{e^{z_i/T}}{\sum_j e^{z_j/T} }$ where $i,j$ index the output units. The introduction of $T$ allows more information of non-targets to be distilled. For example, a training sample with the hard target [1 0 0] does not involve any rank information for the second and third class; with the soft targets, e.g., [0.8, 0.15, 0.5], the rank information of the second and third class is reflected. Additionally, with a large $T$ applied, the target is even softer, e.g, [0.6, 0.25, 0.15], which allows the non-target classes to be more prominent in the training. Note that the additional rank information on the non-target classes is not available in the original target, but is distilled from the teacher model. Additionally, a larger $T$ boosts information of non-target classes but at the same time reduces information of target classes. If $T$ is very large, the soft target falls back to a uniform distribution and is not informative any more\footnote{This argument should be not confused with the conclusion in~\cite{hinton2014distilling} where it was found that when $T$ is also applied to the child net, a large $T$ is equal to logit matching. The assumption of this equivalence is that $T$ is large compared to the magnitude of the logit values, but not infinitely large. In fact, if $T$ is very large, the gradient will approach zero so no knowledge is distilled from the teacher model.}. Therefore, $T$ controls how the knowledge is distilled from the teacher model and hence needs to be set appropriately according to the task in hand.

\subsection{Dark knowledge for complex model training}

Dark knowledge, in the form of soft targets, can be used not only for boosting simple models, but also for training complex models. We argue that training with soft targets offers at least two advantages: (1) it provides more information for model training and (2) it makes the training more reliable. These two advantages are particularly important for training complex models, especially when the training data is limited.

Firstly, soft targets offer probabilistic class labels which are not so `definite' as hard targets. On one hand, this matches the real situation where uncertainty always exists in classification tasks. For example, in speech recognition, it is often difficult to identify the phone class of a frame due to the effect of co-articulation. On the other hand, this uncertainty involves rich (but less discriminative) information within a single example. For example, the uncertainty in phone classes indicates phones are similar to each other and easy to get confused. Making use of this information in the form of soft targets (posterior probabilities) helps improve statistical strength of all phones in a collaborative way, and therefore is particularly helpful for phones with little training data.

Secondly, soft targets blur the decision boundary of classes, which offers a smooth training. The smoothness associated with soft targets has been noticed in~\cite{hinton2014distilling}, which states that soft targets result in less variance in the gradient between training samples. This can be easily verified by looking at the gradients backpropagated to the logit layer, which is $t_i-y_i$ for the $i$-th logit, where $t_i$ is the target and $y_i$ is the output of the child model in training. The accumulated variance is given by:
\[
Var(t) = \sum_i \{ \mathbf{E}_x (t_i - y_i)^2 - (\mathbf{E}_x t_i - \mathbf{E}_x y_i)^2\}
\]
\noindent where the expectation $\mathbf{E}_x$ is conducted on the training data $x$. If we assume that $\mathbf{E}_x t_i$ is identical for soft and hard targets (which is reasonable if the teacher model is well trained on the same data), then the variance is given by:
\begin{eqnarray}
\nonumber
Var (t) &=& \sum_i \mathbf{E}_x (t_i - y_i)^2 + const
\end{eqnarray}
\noindent where $const$ is a constant term. If we assume that the child model can well learn the teacher model, the gradient variance approaches to zero with soft targets, which is impossible with hard targets even if when the training has converged.

The reduced gradient variance is highly desirable when training deep and complex models such as RNNs. We argue that it can  mitigate the risk of gradient vanishing and explosion that is well known to hinder RNN training, leading to a more reliable training.

\subsection{Regularization view}

It has been known that including both soft and hard targets improves performance with appropriate setting of a weight factor to balance their relative contributions~\cite{hinton2014distilling}. This can be formulated as a regularized training problem, with the objective function given by:

\begin{eqnarray}
\nonumber
\mathscr{L} (\theta) &=& \alpha \mathscr{L}_H (\theta) +  \mathscr{L}_S (\theta)\\
\nonumber
                     &=& \sum_i  \sum_j (\alpha t_{ij} +  p_{ij}) ln \{y_{ij} (\theta)\}
\end{eqnarray}
\noindent where $\theta$ represents the parameters of the model, $\mathscr{L}_H (\theta)$ and  $\mathscr{L}_S (\theta)$ are the cost associated with the hard and soft targets respectively, and $\alpha$ is the weight factor. Additionally, $t_{ij}$ and $p_{ij}$ are the hard and soft targets for the $i$-th sample on the $j$-th class, respectively.  Note that $\mathscr{L}_H (\theta)$ is the objective function of the conventional supervised training, and so $\mathscr{L}_S (\theta)$ plays a role of regularization. The effect of the regularization term is to force the model under training (child model) to mimic the teacher model, a way of knowledge transfer. In this study, a DNN model is used as the teacher model to regularize the training of an RNN.  With this regularization, the RNN training looks for optima which produce similar targets as the DNN does, so the risk of over-fitting and under-fitting can be largely reduced.

\subsection{Pre-training view}

Instead of training the model with soft and hard targets altogether, we can first train a reasonable model with soft targets, and then refine the model with hard targets. By this way, the transfer learning plays the role of pre-training, and the conventional supervised training plays the role of fine-tuning. The rationale is that the soft targets results in a reliable training so can be used to conduct model initialization. However, since the information involved in soft targets is less discriminative, refinement with hard targets tends to be helpful. This can be informally interpreted as teaching the model with less but important discriminative information firstly, and once the model is strong enough, more discriminative information can be learned.

This leads to a new pre-training strategy based on dark knowledge transfer. In the conventional pre-training approaches based on either restricted Boltzmann machine (RBM)~\cite{hinton2006reducing} or auto-encoder (AE)~\cite{bengio2007greedy}, simple models are trained and stacked to construct complex models. The dark knowledge pre-training functions in a different way: it makes a complex model trainable by using less discriminative information (soft targets), while the model structure does not change. This approach possesses several advantages: (1) it is totally supervised and so more task-oriented; (2) it pre-trains the model as a whole, instead of layer by layer, so tends to be fast; (3) it can be used to pre-train any complex models for which the layer structure is not clear, such as the RNN model that we focus on in this paper.

The pre-training view is related to the curriculum training method discussed in~\cite{romero2014fitnets}, where training samples that are easy to learn are firstly selected to train the model, while more difficult ones are selected later when the model has been fairly strong. In the dark knowledge pre-training, the soft targets can be regarded as easy samples for pre-training, and hard targets as difficult samples for fine-tuning.

Interestingly, the regularization view and the pre-training view are closely related. The pre-training is essentially a regularization that places the model to some location in the parameter space where good local minima can be easily reached. This relationship between regularization and pre-training has been discussed in the context of DNN training~\cite{erhan2010does}.

\section{Experiments}
\label{sec:exp}

To verify the proposed method, we use it to train RNN acoustic models for an ASR task which is known to be difficult. Note that all the RNNs we mention in this section are indeed LSTMs. The experiments are conducted on the Aurora4 database in noisy conditions, and the data profile is largely standard: $7137$ utterances for model training,  $4620$ utterances for development and $4620$ utterances for testing. The Kaldi toolkit\cite{Povey_ASRU2011} is used to conduct the model training and performance evaluation, and the process largely follows the Aurora4 s5 recipe for GPU-based DNN training. Specifically, the training starts from constructing a system based on Gaussian mixture models (GMM) with the standard $13$-dimensional MFCC features plus the first and second order derivatives. A DNN system is then trained with the alignment provided by the GMM system. The feature used for the DNN system is the $40$-dimensional Fbanks. A symmetric $11$-frame window is applied to concatenate neighboring frames, and an LDA transform is used to reduce the feature dimension to $200$, which forms the DNN input. The DNN architecture involves $4$ hidden layers and each layer consists of $2048$ units. The output layer is composed of $2008$ units, equal to the total number of Gaussian mixtures in the GMM system. The cross entropy is used as the training criterion, and the stochastic gradient descendent (SGD) algorithm is employed to perform the training.

In the dark knowledge transfer learning, the trained DNN model is used as the teacher model to generate soft targets for the RNN training. The RNN architecture involves 2 layers of LSTMs with 800 cells per layer. The unidirectional LSTM has a recurrent projection layer as in~\cite{sak2014long} while the non-recurrent one is discarded. The input features are the $40$-dimensional Fbanks, and the output units correspond to the Gaussian mixtures as in the DNN.
The RNN is trained with $4$ streams and each stream contains $20$ continuous frames.
The momentum is empirically set to $0.9$, and the starting learning rate is set to $0.0001$ by default.

The experimental results are reported in Table~\ref{tab:res}. The performance is evaluated in terms of two criteria: the frame accuracy (FA) and the word error rate (WER). While FA is more related to the training criterion (cross entropy), WER is more important for speech recognition. In Table~\ref{tab:res}, the FAs are reported on both the training set (TR FA) and the cross validation set (CV FA), and the WER is reported on the test set.

In Table~\ref{tab:res}, RNN-0 is the RNN baseline trained with hard targets. RNN-T1 and RNN-T2 are trained with dark knowledge transfer, where the temperature $T$ is set to $1$ and $2$ respectively. For each dark knowledge transfer model, the soft targets are employed in three ways: in the `soft' way, only soft targets are used in RNN training; in the `reg.' way, the soft and hard targets are used together, and the soft targets play the role of regularization, where the gradients of the soft's are scaled up with $T^2$~\cite{hinton2014distilling}. In the `pretrain' way, the soft targets and the hard targets are used sequentially, and the soft targets play the role of pre-training. The weight factor in the regularization approach is empirically set to $0.5$.

\begin{table}[!htb]
\centering
\begin{tabular}{l|c|c|c|c}
\hline
                    & Targets              & FA\% & FA\%   & WER\% \\
                    &                      & TR &CV   &  \\
\hline
DNN                 & Hard         	       & 63.0 & 45.2   & 11.40 \\
RNN-0               & Hard                 & 67.3 & 51.9   & 13.57  \\
\hline
RNN-T1 (soft)        & Soft                & 59.4 & 49.9   & 11.46 \\
RNN-T1 (reg.)        & Soft + Hard         & 67.5 & 53.7   & 10.84 \\
RNN-T1 (pretrain)    & Soft, Hard         & 65.5 & 54.2   & 10.71 \\
\hline
RNN-T2  (soft)       & Soft                &58.2  & 49.5   & 11.32   \\
RNN-T2  (reg.)       & Soft + Hard         &65.8  & 53.3   & 10.88	 \\
RNN-T2  (pretrain)   & Soft, Hard         &64.6  & 54.1   & 10.57	 \\
\hline
\end{tabular}
\caption{Results with Different Models and Training Methods}
\label{tab:res}
\end{table}

It can be observed that the RNN baseline (RNN-0) can not beat the DNN baseline in terms of WER, although much effort has been devoted to calibrate the training process, including various trials on different learning rates and momentum values. This is consistent with the results published with the Kaldi recipe. Note that this does not mean RNNs are inferior to DNNs. From the FA results, it is clear that the RNN model leads to better quality in terms of the training objective. Unfortunately, this advantage is not propagated to WER on the test set. Additionally, the results shown here can not be interpreted as that RNNs are not suitable for ASR (in terms of WER). In fact several researchers have reported better WERs with RNNs, e.g.,~\cite{graves2014towards}. Our results just say that with the Aurora4 database, the RNN with the \emph{basic} training method does not generalize well in terms of WER, although it works well in terms of the training criterion.

This problem can be largely solved by the dark knowledge transfer learning, as demonstrated by the results of the RNN-T1 and RNN-T2 systems. It can be seen that with the soft targets only, the RNN system obtains equal (T=1) or even better (T=2) performance in comparison with the DNN baseline, which means that the knowledge embedded in the DNN model has been transferred to the RNN model, and the knowledge can be arranged in a better form within the RNN structure. Paying attention to the FA results, it can be seen that the knowledge transfer learning does not improve accuracy on the training set, but leads to better or close FAs on the CV set compared to the DNN and RNN baseline. This indicates that transfer learning with soft targets sacrifices the FA performance on the training set a little, but leads to better generalization on the CV set. Additionally, the advantage on WER indicates that the generalization is improved not only in the sense of data sets, but also in the sense of evaluation metrics.

When combining soft and hard targets, either in the way of regularization or pre-training, the performance in terms of both FA and WER is improved. This confirms the hypothesis that the knowledge transfer learning does play roles of regularization and pre-training. Note that in all these cases, the FA results on the training set are lower than that of the RNN baseline, which confirms that the advantage of the knowledge transform learning resides in improving generalizability of the resultant model. When comparing the two dark knowledge RNN systems with different temperatures $T$, we see T=2 leads to little worse FAs on the training and CV set, but slightly better WERs. This confirms that a higher temperature generates a smoother direction and leads to better generalization.

\section{Conclusion}
\label{sec:con}

We proposed a novel RNN training method based on dark knowledge transfer learning. The experimental results on the ASR task demonstrated that knowledge learned by simple models can be effectively used to guide the training of complex models. This knowledge can be used either as a regularization or for pre-training, and both approaches can lead to models that are more generalizable, a desired property for complex models. The future work involves applying this technique to more complex models that are difficult to train with conventional approaches, for example deep RNNs. Knowledge transfer between heterogeneous models is under investigation as well, e.g., between probabilistic models and neural models.

\newpage
\bibliographystyle{IEEEtran}

\bibliography{dark}

\begin{thebibliography}{10}
\providecommand{\url}[1]{#1}
\csname url@samestyle\endcsname
\providecommand{\newblock}{\relax}
\providecommand{\bibinfo}[2]{#2}
\providecommand{\BIBentrySTDinterwordspacing}{\spaceskip=0pt\relax}
\providecommand{\BIBentryALTinterwordstretchfactor}{4}
\providecommand{\BIBentryALTinterwordspacing}{\spaceskip=\fontdimen2\font plus
\BIBentryALTinterwordstretchfactor\fontdimen3\font minus
  \fontdimen4\font\relax}
\providecommand{\BIBforeignlanguage}[2]{{%
\expandafter\ifx\csname l@#1\endcsname\relax
\typeout{** WARNING: IEEEtran.bst: No hyphenation pattern has been}%
\typeout{** loaded for the language `#1'. Using the pattern for}%
\typeout{** the default language instead.}%
\else
\language=\csname l@#1\endcsname
\fi
#2}}
\providecommand{\BIBdecl}{\relax}
\BIBdecl

\bibitem{deng2014}
\BIBentryALTinterwordspacing
L.~Deng and D.~Yu, ``Deep learning: Methods and applications,''
  \emph{Foundations and Trends in Signal Processing}, vol.~7, no. 3-4, pp.
  197--387, 2013. [Online]. Available:
  \url{http://dx.doi.org/10.1561/2000000039}
\BIBentrySTDinterwordspacing

\bibitem{graves2013speech}
A.~Graves, A.-R. Mohamed, and G.~Hinton, ``Speech recognition with deep
  recurrent neural networks,'' in \emph{Proceedings of IEEE International
  Conference on Acoustics, Speech and Signal Processing (ICASSP)}.\hskip 1em
  plus 0.5em minus 0.4em\relax IEEE, 2013, pp. 6645--6649.

\bibitem{graves2014towards}
A.~Graves and N.~Jaitly, ``Towards end-to-end speech recognition with recurrent
  neural networks,'' in \emph{Proceedings of the 31st International Conference
  on Machine Learning (ICML-14)}, 2014, pp. 1764--1772.

\bibitem{sak2014long}
H.~Sak, A.~Senior, and F.~Beaufays, ``Long short-term memory recurrent neural
  network architectures for large scale acoustic modeling,'' in
  \emph{Proceedings of the Annual Conference of International Speech
  Communication Association (INTERSPEECH)}, 2014.

\bibitem{rumelhart1988learning}
\BIBentryALTinterwordspacing
D.~E. Rumelhart, G.~E. Hinton, and R.~J. Williams, ``Learning representations
  by back-propagating errors,'' \emph{Nature}, vol. 323, no. 6088, pp.
  533--536, 1986, 10.1038/323533a0. [Online]. Available:
  \url{http://dx.doi.org/10.1038/323533a0}
\BIBentrySTDinterwordspacing

\bibitem{bengio1994learning}
Y.~Bengio, P.~Simard, and P.~Frasconi, ``Learning long-term dependencies with
  gradient descent is difficult,'' \emph{Neural Networks, IEEE Transactions
  on}, vol.~5, no.~2, pp. 157--166, 1994.

\bibitem{hochreiter1997long}
S.~Hochreiter and J.~Schmidhuber, ``Long short-term memory,'' \emph{Neural
  computation}, vol.~9, no.~8, pp. 1735--1780, 1997.

\bibitem{graves2005framewise}
A.~Graves and J.~Schmidhuber, ``Framewise phoneme classification with
  bidirectional lstm and other neural network architectures,'' \emph{Neural
  Networks}, vol.~18, no.~5, pp. 602--610, 2005.

\bibitem{jaeger2004harnessing}
H.~Jaeger and H.~Haas, ``Harnessing nonlinearity: Predicting chaotic systems
  and saving energy in wireless communication,'' \emph{Science}, vol. 304, no.
  5667, pp. 78--80, 2004.

\bibitem{martens2010deep}
J.~Martens, ``Deep learning via hessian-free optimization,'' in
  \emph{Proceedings of the 27th International Conference on Machine Learning
  (ICML-10)}, 2010, pp. 735--742.

\bibitem{martens2011learning}
J.~Martens and I.~Sutskever, ``Learning recurrent neural networks with
  hessian-free optimization,'' in \emph{Proceedings of the 28th International
  Conference on Machine Learning (ICML-11)}, 2011, pp. 1033--1040.

\bibitem{sutskever2013importance}
I.~Sutskever, J.~Martens, G.~Dahl, and G.~Hinton, ``On the importance of
  initialization and momentum in deep learning,'' in \emph{Proceedings of the
  30th International Conference on Machine Learning (ICML-13)}, 2013, pp.
  1139--1147.

\bibitem{ba2014deep}
J.~Ba and R.~Caruana, ``Do deep nets really need to be deep?'' in
  \emph{Advances in Neural Information Processing Systems}, 2014, pp.
  2654--2662.

\bibitem{hinton2014distilling}
G.~E. Hinton, O.~Vinyals, and J.~Dean, ``Distilling the knowledge in a neural
  network,'' in \emph{NIPS 2014 Deep Learning Workshop}, 2014.

\bibitem{bucilu¨£2006model}
C.~Bucilu¨£, R.~Caruana, and A.~Niculescu-Mizil, ``Model compression,'' in
  \emph{Proceedings of the 12th ACM SIGKDD international conference on
  Knowledge discovery and data mining}.\hskip 1em plus 0.5em minus 0.4em\relax
  ACM, 2006, pp. 535--541.

\bibitem{li2014learning}
J.~Li, R.~Zhao, J.-T. Huang, and Y.~Gong, ``Learning small-size {DNN} with
  output-distribution-based criteria,'' in \emph{Proceedings of the Annual
  Conference of International Speech Communication Association (INTERSPEECH)},
  2014.

\bibitem{chan2015transferring}
W.~Chan, N.~R. Ke, and I.~Lane, ``Transferring knowledge from a {RNN} to a
  {DNN},'' \emph{arXiv preprint arXiv:1504.01483}, 2015.

\bibitem{hinton2006reducing}
G.~E. Hinton and R.~R. Salakhutdinov, ``Reducing the dimensionality of data
  with neural networks,'' \emph{Science}, vol. 313, no. 5786, pp. 504--507,
  2006.

\bibitem{bengio2007greedy}
Y.~Bengio, P.~Lamblin, D.~Popovici, H.~Larochelle \emph{et~al.}, ``Greedy
  layer-wise training of deep networks,'' \emph{Advances in neural information
  processing systems}, vol.~19, p. 153, 2007.

\bibitem{romero2014fitnets}
A.~Romero, N.~Ballas, S.~E. Kahou, A.~Chassang, C.~Gatta, and Y.~Bengio,
  ``Fitnets: Hints for thin deep nets,'' \emph{arXiv preprint arXiv:1412.6550},
  2014.

\bibitem{erhan2010does}
D.~Erhan, Y.~Bengio, A.~Courville, P.-A. Manzagol, P.~Vincent, and S.~Bengio,
  ``Why does unsupervised pre-training help deep learning?'' \emph{The Journal
  of Machine Learning Research}, vol.~11, pp. 625--660, 2010.

\bibitem{Povey_ASRU2011}
D.~Povey, A.~Ghoshal, G.~Boulianne, L.~Burget, O.~Glembek, N.~Goel,
  M.~Hannemann, P.~Motlicek, Y.~Qian, P.~Schwarz, J.~Silovsky, G.~Stemmer, and
  K.~Vesely, ``The kaldi speech recognition toolkit,'' in \emph{IEEE 2011
  Workshop on Automatic Speech Recognition and Understanding}.\hskip 1em plus
  0.5em minus 0.4em\relax IEEE Signal Processing Society, Dec. 2011, iEEE
  Catalog No.: CFP11SRW-USB.

\end{thebibliography}

\end{document}